\documentclass[10pt,twocolumn,letterpaper]{article}

\usepackage[pagenumbers]{cvpr} 

\usepackage{url}
\usepackage{graphicx}

\usepackage{caption}
\usepackage{booktabs}
\usepackage{algorithm}
\usepackage{algorithmic}
\usepackage{siunitx}
\usepackage{wrapfig}
\usepackage{graphicx}
\usepackage[most]{tcolorbox} 
\definecolor{cvprblue}{rgb}{0.21,0.49,0.74}
\usepackage[pagebackref,breaklinks,colorlinks,allcolors=cvprblue]{hyperref}

\def\paperID{14617} 
\def\confName{CVPR}
\def\confYear{2026}

\title{Iterative Refinement Improves Compositional Image Generation}

\author{Shantanu Jaiswal$^{1}$ \quad
Mihir Prabhudesai$^{1}$ \quad
Nikash Bhardwaj$^{1}$ \quad
Zheyang Qin$^{1}$ \\
Amir Zadeh$^{2}$ \quad
Chuan Li$^{2}$ \quad
Katerina Fragkiadaki$^{1}$ \quad
Deepak Pathak$^{1}$ \\
\\
$^{1}$Carnegie Mellon University \quad
$^{2}$Lambda AI
}

\begin{document}

\twocolumn[{%
\vspace{-2.9em}
    \maketitle
    \begin{center}
        \vspace{-2.5em}
        {\url{https://iterative-img-gen.github.io/}}
        \vspace{1em}
    \end{center}
    
    \begin{center}
        \centering
        \vspace{-8pt}
        \includegraphics[width=\linewidth,trim=391 30 391 128,clip]{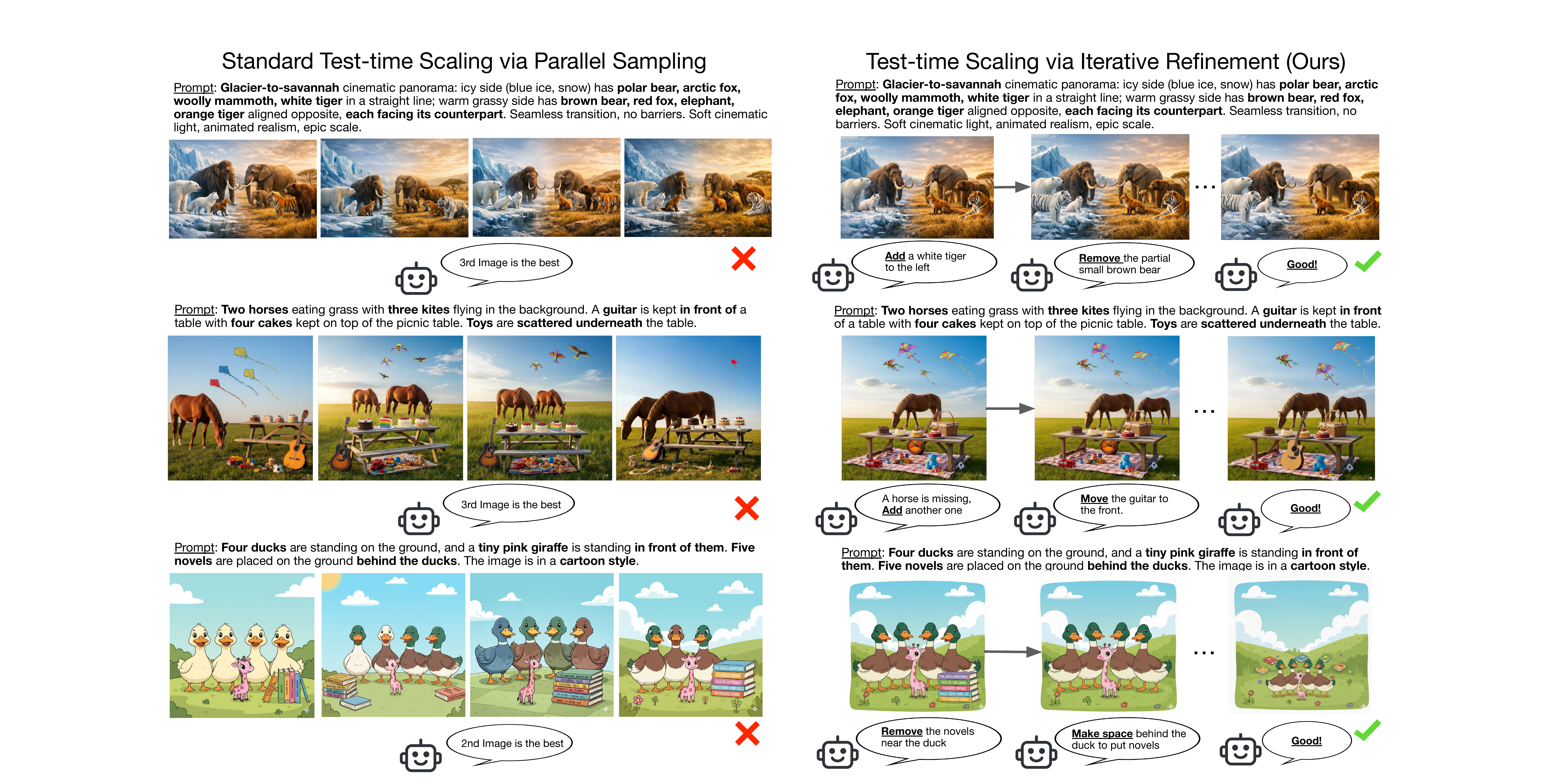}
        \vspace{-13.5pt}
        \captionof{figure}{Iterative refinement during inference time enables high fidelity generation of complex prompts on which traditional inference-time scaling strategies such as parallel sampling can fail to generate a fully accurate image even at high num. of samples as shown above.
        }
        \label{fig:main_results}
    \end{center}
}]

\begin{abstract}
Text-to-image (T2I) models have achieved remarkable progress, yet they continue to struggle with complex prompts that require simultaneously handling multiple objects, relations, and attributes. Existing inference-time strategies, such as parallel sampling with verifiers or simply increasing denoising steps, can improve prompt alignment but remain inadequate for richly compositional settings where many constraints must be satisfied. Inspired by the success of chain-of-thought reasoning in large language models, we propose an iterative test-time strategy in which a T2I model progressively refines its generations across multiple steps, guided by feedback from a vision-language model as the critic in the loop. Our approach is simple, requires no external tools or priors, and can be flexibly applied to a wide range of image generators and vision-language models. Empirically, we demonstrate consistent gains on image generation across benchmarks: a $16.9\%$ improvement in all-correct rate on ConceptMix (k=7), a $13.8\%$ improvement on T2I-CompBench (3D-Spatial category) and a $12.5\%$ improvement on Visual Jenga scene decomposition compared to compute-matched parallel sampling. Beyond quantitative gains, iterative refinement produces more faithful generations by decomposing complex prompts into sequential corrections, with human evaluators preferring our method 58.7\% of the time over 41.3\% for the parallel baseline. Together, these findings highlight iterative self-correction as a broadly applicable principle for compositional image generation.

\end{abstract}    
\section{Introduction}
\label{sec:intro}

Large language models (LLMs) have achieved remarkable progress in recent years, as a result of simply scaling test-time compute \cite{wei2022chain,brown2024large,snell2024scaling}. A particularly influential development has been the use of chain-of-thought (CoT) prompting, where models are instructed to “think step by step” \cite{wei2022chain,kojima2022large}. Despite its simplicity, this strategy enables models to exhibit sophisticated behaviors such as self-correction, error checking, and iterative refinement, ultimately leading to significant gains on reasoning-intensive benchmarks. These behaviors highlight the potential of LLMs not only as static predictors but as systems that can actively refine their outputs through structured intermediate reasoning.

The success of CoT reasoning in LLMs is closely tied to their pre-training data. During training, LLMs are exposed to large volumes of text that naturally contain traces of human step-by-step reasoning -- mathematical derivations, logical arguments, and instructional writing. This supervision on the internet implicitly provides the prior that chain-of-thought prompting later exploits, enabling the model to perform multi-step reasoning. By contrast, text-to-image (T2I) models are trained on large-scale datasets of image–caption pairs that lack such structured reasoning traces. As a result, these models do not inherently develop capabilities like self-correction or iterative refinement, instead rely on one-shot generation strategies that limit their robustness in complex settings.
\begin{figure}[t]
  \begin{center}
    \includegraphics[width=1.03\linewidth]{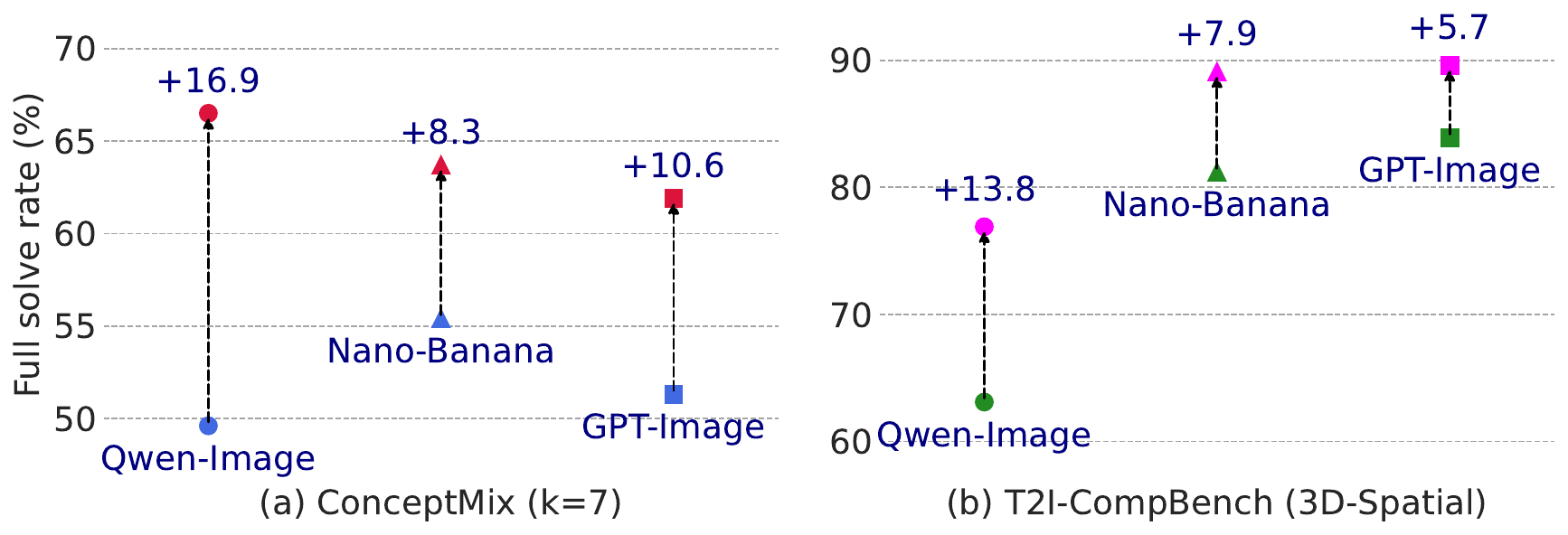}
  \end{center}
  \caption{Our iterative inference-time strategy achieves strong benefits over computation-matched parallel inference time scaling on multiple state-of-art image generation models.}
  \label{fig:figure_intro}
\end{figure}

In this work, we investigate how can we enable self-correction in T2I models. Our central idea is to leverage complementary modules that together mimic the iterative reasoning process observed in LLMs. Concretely, our framework integrates four components:  (i) a text-to-image (T2I) model to generate an initial image, (ii) a vision-language model (VLM) critic to propose corrections by comparing the generated image with target prompt, (iii) an image editor to apply suggested edits, and (iv) a verifier to evaluate alignment between final image and target prompt. This pipeline allows the model to iteratively refine its outputs rather than relying solely on a single forward pass.

We compare our approach against the widely adopted strategy of parallel sampling \cite{ma2025inference,zhang2025inference}, where multiple images are generated independently and the best one is selected using a verifier. While parallel sampling increases diversity, it does not fundamentally change the underlying generation process, nor does it allow the model to revise or build upon earlier outputs. As a result, it struggles with complex compositional prompts. For example, consider a prompt requiring dozens of concept bindings: if the model’s attention heads cannot jointly resolve all bindings in a single forward pass, the pass@k will remain near zero regardless of how many samples are drawn.

In contrast, our approach explicitly reuses intermediate generations and progressively improves them through guided corrections. This factorization allows the model to handle only a subset of bindings at each step, compounding previously resolved components over time. Such sequential, step-by-step refinement—analogous to chain-of-thought reasoning— is crucial for reliably generating highly compositional images.

Figure \ref{fig:main_results}, highlights the capability of our approach to generate complex compositional prompts. Given the caption on top, parallel sampling simply is unable to build on top of the previous steps thus being unsuccessful even after 4 passes through the generative model. In contrast, iterative refinement successfully generates the final image, while using the same amount of compute. Quantitatively in Figure \ref{fig:figure_intro}, we demonstrate that this leads to consistent performance improvements: our approach achieves a 16.9\% higher all-correct rate on ConceptMix \cite{wu2024conceptmix} (for concept binding=7) and a 13.8\% gain on T2I-Bench 3D Spatial category \cite{huang2023t2i} relative to compute-matched parallel sampling.

An alternate family of methods—such as GenArtist \cite{wang2024genartist} and CompAgent \cite{wang2024divide}—also performs sequential sampling by building on top of previous generations. However, these approaches rely on a large toolbox of auxiliary modules (e.g., layout-to-image models, bounding-box detectors, dragging tools, and object-removal systems). Because these toolchains evolve at different rates and often lag behind foundation model capabilities, the overall pipeline becomes brittle: errors from individual tools can accumulate rather than help the generation process for complex prompts. Other methods such as RPG \cite{yang2024mastering} similarly show gains via increased test-time compute, but still depend on complex region-wise priors and bespoke pipelines not readily applicable to black-box foundation models.

In contrast, with recent advances in VLMs and modern image-editing models, we find that many of these specialized tool-based pipelines are no longer necessary for effective test-time scaling. Across all benchmarks, simply combining a strong VLM critic feedback generator with a standard image-editing model is sufficient to achieve state-of-the-art compositional image generation -- without relying on heavy tool stacks or model-specific training and engineering pipelines. As shown in Figure~\ref{fig:method_comparison}, methods such as GenArtist and RPG under-perform substantially in highly compositional settings, whereas our approach delivers a consistent  $\sim 9{+}\%$ point improvement under matched compute. Further, our framework naturally extends to the recent Visual Jenga  scene decomposition task \cite{bhattad2025visual} as detailed in sec.~\ref{sec:visual-jenga}. 

Our findings further suggest that self-correction---long recognized as a key ingredient in LLM reasoning---also serves as a powerful inductive principle for generative vision models. Introducing a simple and general refinement pipeline enables behaviors traditionally associated with language models to naturally transfer into image generation, yielding tangible performance gains. More broadly, this work points to the promise of designing generative systems that not only produce outputs but also critique and improve upon them, moving towards a more unified view of reasoning across modalities.

\section{Related Work}
\label{related_work}

\textbf{Text-to-Image Inference-Time Strategies.}  
Recent advances in text-to-image (T2I) generation have demonstrated impressive capabilities in producing high-quality and diverse images from natural language prompts \cite{goodfellow2016deep,hinton2006fast,bengio2007scaling}. However, complex prompts with multiple objects, relations, and fine-grained attributes remain challenging. Inference-time strategies such as classifier-free guidance \cite{ho2022classifier}, parallel sampling \cite{feng2023layoutgpt,chefer2023attend}, and grounding-based methods \cite{li2023gligen,lian2023llm} improve prompt fidelity but often fail to scale to richly compositional prompts. Iterative refinement methods, including SDEdit \cite{meng2021sdedit}, InstructPix2Pix \cite{brooks2023instructpix2pix}, and IterComp\cite{zhang2024itercomp} attempt to progressively improve image alignment with prompts by using multiple generation steps and feedback mechanisms. Human-preference-guided evaluation and optimization, as in \cite{lee2023aligning,wu2023human,kirstain2023pick}, further highlight the importance of incorporating adaptive guidance at inference time. T2I models \cite{openai2023dalle3, wu2025qwen, openai2025gptimage1, blackforest2024flux}  and compositional methods such as IterComp~\cite{zhang2024itercomp}, RPG~\cite{yang2024mastering}, GenArtist~\cite{wang2024genartist}, PARM~\cite{zhang2025let}, LLM Diffusion~\cite{lian2023llm}  and CompAgent~\cite{wang2024divide} are related to our method, but either make use of tool-calling, regional generation priors or reinforcement learning objectives to improve compositionality. In contrast, our method is a training free method with simply a VLM-critic utilized in loop with an image generation and editing model, and empirically demonstrates stronger performance benefits across different T2I model families.    

\textbf{Chain-of-Thought Reasoning in Large Language Models.}  
Chain-of-thought (CoT) prompting has been shown to elicit multi-step reasoning and improve performance on complex language tasks \cite{wei2022chain,wang2022self,yao2023tree}. Iterative and self-refinement approaches \cite{madaan2023self} further demonstrate that models benefit from decomposing a problem into sequential reasoning steps with feedback loops. Drawing inspiration from these strategies, our method applies a similar iterative reasoning paradigm to T2I generation: the critic functions analogously to a CoT process, first evaluating candidate generations and then issuing targeted refinement prompts, enabling high-fidelity compositional image synthesis.

\section{Method}
\label{method}

\begin{figure*}[t]
    \centering
    \includegraphics[width=0.9\linewidth]{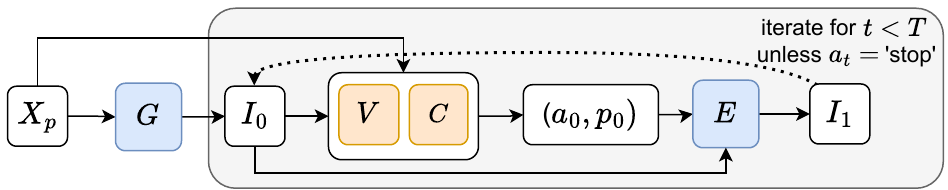}
    \caption{Given a complex text prompt $X_p$, a generator $G$ produces an initial image $I_0$. A test-time verifier $V$ and critic $C$, conditioned on $X_p$, output an action--sub-prompt pair $(a_t, p_t)$. The previous image $I_{t-1}$ and sub-prompt $p_t$ are fed to an editor $E$ to yield the next image $I_t$. This process repeats under an inference-time budget $B$, allocated as maximum $T$ iterative rounds over $M$ parallel streams, until a \texttt{STOP} action is emitted or $B$ ($=T\times M$) is exhausted.}
    \label{fig:method}
\end{figure*}

Given a complex text prompt $P$, our goal is to generate an image $I$ that faithfully captures all entailed entities and compositions. We adopt an \emph{iterative inference-time refinement} scheme in which a generator progressively improves its outputs under critic guidance, subject to an inference-time computational budget $B$ that we allocate as $T$ refinement rounds across $M$ parallel streams.

\textbf{Setup.} Let $G$ denote a text-to-image generator, $E$ a image-to-image editor, $V$ a verifier that scores alignment between a candidate image and prompt $P$, and $C$ a critic that outputs (i) a refinement sub-prompt $p_t$ to guide subsequent updates, and (ii) an \emph{action} $a_t \in \{\texttt{STOP}, \texttt{BACKTRACK}, \texttt{RESTART}, \texttt{CONTINUE}\}$ indicating how the refinement should be applied. We assume an inference-time budget $B$, allocated into $T$ refinement rounds over $M$ parallel streams (i.e., $B = T \times M$ unit refinement operations). This parameterization exposes a controllable depth--breadth trade-off, where each unit corresponds to a single call to the text-to-image generator $G$ or the image-to-image editor $E$. The action space of the critic $C$ is as follows:  
\begin{itemize}
    \item \texttt{STOP}: terminate the process upon completion satisfaction and return the current image.  
    \item \texttt{BACKTRACK}: revert to the previous generation $I_{t-1}$ and refine it using the new sub-prompt $p_t$.  
    \item \texttt{RESTART}: discard the current trajectory and regenerate from scratch conditioned on $P$ and new sub-prompt $p_t$.  
    \item \texttt{CONTINUE}: refine the current best candidate $I_t^\ast$ directly using new sub-prompt $p_t$.  
\end{itemize}
Unlike standard single-shot or naive parallel sampling, our method unfolds over $T$ refinement rounds and $M$ parallel streams under a fixed budget $B$, where intermediate generations are evaluated, critiqued, and selectively improved.

\textbf{Iterative refinement over parallel streams with critic feedback.} At $t=0$, each parallel stream $m$ initializes a candidate $I_0^m \leftarrow G(P)$, where $P$ is the user image prompt. At iteration $t$, the verifier scores $s_t^m \leftarrow V(I_t^m, P)$ and the critic proposes $(a_t^m, p_t^m) \leftarrow C(I_t^m, P)$. Depending on $a_t^m$, we either stop, backtrack to $I_{t-1}^m$, restart from $G(P, p_t^m)$, or continue editing $I_t^m$ via $E$. The process terminates when a \texttt{STOP} action is emitted for all streams or when the budget $B$ (parameterized by $T$ and $M$) is exhausted.

This procedure enables T2I models to decompose complex compositional prompts into a sequence of refinement steps, akin to chain-of-thought reasoning in LLMs. The verifier ensures consistent prompt alignment, while critic provides targeted feedback to correct systematic errors.

Note that the verifier $V$ is \emph{not} an oracle or benchmark evaluator; rather, it is a lightweight VLM used solely to provide automatic test-time guidance and improvement signals during refinement. We outline the pseudo-code of our iterative refinement technique in Algorithm Block \ref{alg:iterative}.

\begin{algorithm}
\caption{Iterative Image Refinement over Parallel Streams with Critic Feedback}
\label{alg:iterative}
\begin{algorithmic}[1]
\STATE \textbf{Input:} Prompt $P$, generator $G$, non-oracle test-time verifier $V$, critic $C$, parallel streams $M$, max iterative rounds $T$; inference-time budget $B$ allocated as $T \times M$ unit updates
\STATE Initialize $\{I_0^m\}_{m=1}^M \leftarrow \{G(P)\}_{m=1}^M$
\FOR{$t = 1$ to $T$}
    \FOR{$m = 1$ to $M$ \textbf{ in parallel}}
        \STATE Score candidate $s_t^m \leftarrow V(I_t^m, P)$
        \STATE Critic outputs $(a_t^m, p_t^m) \leftarrow C(I_t^m, P)$
        \IF{$a_t^m =$ \texttt{STOP}}
            \STATE Mark stream $m$ as complete
        \ELSIF{$a_t^m =$ \texttt{BACKTRACK}}
            \STATE Revert to $I_{t-1}^m$ and refine: $I_{t+1}^m \leftarrow E(I_{t-1}^m, p_t^m)$
        \ELSIF{$a_t^m =$ \texttt{RESTART}}
            \STATE Reset $I_{t+1}^m \leftarrow G(P, p_t^m)$
        \ELSIF{$a_t^m =$ \texttt{CONTINUE}}
            \STATE Update $I_{t+1}^m \leftarrow E(I_t^m, p_t^m)$
        \ENDIF
    \ENDFOR
    \STATE Select best across all streams: $I_t^\ast \leftarrow \arg\max_m s_t^m$
    \IF{all streams stopped}
        \STATE \textbf{return} $I_t^\ast$
    \ENDIF
\ENDFOR
\STATE \textbf{return} $I_T^\ast$
\end{algorithmic}
\label{alg:iterative}
\end{algorithm}

\section{Experiments}

\begin{table*}[t]
\centering
\footnotesize
\setlength{\tabcolsep}{3pt}
\begin{tabular}{l|ccccccc|cccccccc}
\toprule
& \multicolumn{7}{c|}{\textbf{ConceptMix full solve rate (\%)}} & \multicolumn{8}{c}{\textbf{T2I-CompBench VLLM (GPT4o)  score (1 to 100)}} \\
\cmidrule(lr){2-8} \cmidrule(lr){9-16}
\textbf{Model} & \textbf{k=1} & \textbf{k=2} & \textbf{k=3} & \textbf{k=4} & \textbf{k=5} & \textbf{k=6} & \textbf{k=7} & \textbf{Spatial} & \textbf{3DSpat} & \textbf{Numer} & \textbf{Shape} & \textbf{Color} & \textbf{Texture} & \textbf{Non-Spat} & \textbf{Cmplex} \\
\midrule
Qwen Parallel & 92.8 & 82.5 & 74.3 & 69.2 & 60.1 & 51.2 & 49.6 & 82.3 & 63.1 & 87.0 & 87.2 & 92.6 & \textbf{96.2} & 92.8 & 93.4 \\
Qwen Iter (ours)           & 96.1 & 91.4 & 87.0 & 82.1 & \textbf{79.6} & 67.4 & 64.3 & 87.4 & \textbf{77.3} & 91.1 & 91.2 & 92.4 & 95.1 & \textbf{94.8} & 94.8 \\
Qwen Iter.+Par. (ours)           & \textbf{96.5} & \textbf{91.7} & \textbf{87.4} & \textbf{82.2} & 78.9 & \textbf{71.8} & \textbf{66.5} & \textbf{89.4} & 76.9 & \textbf{93.3} & \textbf{90.1} & 92.6 & 95.8 & 94.7 & \textbf{95.0} \\
& \textcolor{blue}{+3.7} & \textcolor{blue}{+9.2} & \textcolor{blue}{+13.1} & \textcolor{blue}{+13.0} & \textcolor{blue}{+18.8} & \textcolor{blue}{+20.6} & \textcolor{blue}{+16.9} & \textcolor{blue}{+7.1} & \textcolor{blue}{+13.8} & \textcolor{blue}{+6.3} & \textcolor{blue}{+2.9} & \textcolor{blue}{+0.0} & \textcolor{blue}{-0.4} & \textcolor{blue}{+1.9} & \textcolor{blue}{+1.6} \\
\midrule
Nano-Banana Parallel   & 93.8 & 88.8 & 86.6 & 78.4 & 65.8 & 61.7 & 55.4 & 84.7 & 81.2 & 84.3 & 88.5 & 89.8 & 95.0 & \textbf{96.8} & 91.0 \\
Nano-Banana Iter (ours) & 94.1 & 90.4 & 87.2 & 81.3 & \textbf{73.5} & 64.6 & 63.6 & 90.6 & 87.8 & 93.9 & 89.9 & 89.7 & \textbf{95.1} & 95.8 & \textbf{94.7} \\
Nano-Banana Iter.+Par. (ours) & 93.8 & \textbf{91.0} & \textbf{87.5} & \textbf{82.8} & 71.4 & \textbf{69.8} & \textbf{63.7} & \textbf{91.1} & \textbf{89.1} & \textbf{94.1} & \textbf{88.8} & \textbf{92.1} & 94.8 & 96.7 & 94.5 \\
& \textcolor{blue}{+0.0} & \textcolor{blue}{+2.2} & \textcolor{blue}{+0.9} & \textcolor{blue}{+4.4} & \textcolor{blue}{+5.6} & \textcolor{blue}{+8.1} & \textcolor{blue}{+8.3} & \textcolor{blue}{+6.6} & \textcolor{blue}{+7.9} & \textcolor{blue}{+9.8} & \textcolor{blue}{+0.3} & \textcolor{blue}{+2.3} & \textcolor{blue}{-0.2} & \textcolor{blue}{-0.1} & \textcolor{blue}{+3.5} \\
\midrule
GPT-Image Parallel           & 94.2 & 89.2 & 88.1 & 76.7 & 71.0 & 69.5 & 51.3 & 87.5 & 83.9 & 88.6 & 88.5 & 91.6 & \textbf{92.5} & 95.3 & 92.9 \\
GPT-Image Iterative (ours)            & 96.0 & 91.4 & 90.6 & \textbf{85.4} & 72.0 & 69.6 & 58.9 & 89.6 & \textbf{90.0} & 92.7 & \textbf{92.1} & \textbf{91.9} & 92.0 & \textbf{95.5} & 93.0 \\
GPT-Image Iter.+Par. (ours)            & \textbf{97.7} & \textbf{94.2} & \textbf{91.1} & 84.6 & \textbf{79.5} & \textbf{76.8} & \textbf{61.9} & \textbf{91.0} & 89.6 & \textbf{93.2} & 90.9 & 91.1 & 92.3 & 95.3 & \textbf{93.1} \\
& \textcolor{blue}{+3.5} & \textcolor{blue}{+5.0} & \textcolor{blue}{+3.0} & \textcolor{blue}{+7.9} & \textcolor{blue}{+8.5} & \textcolor{blue}{+7.3} & \textcolor{blue}{+10.6} & \textcolor{blue}{+3.5} & \textcolor{blue}{+5.7} & \textcolor{blue}{+4.6} & \textcolor{blue}{+2.4} & \textcolor{blue}{-0.5} & \textcolor{blue}{-0.2} & \textcolor{blue}{+0.0} & \textcolor{blue}{+0.2} \\
\bottomrule
\end{tabular}
\caption{Performance comparison of parallel sampling, iterative refinement, and combined strategies across three state-of-the-art text-to-image models on ConceptMix \cite{wu2024conceptmix} and T2I-CompBench \cite{huang2023t2i}. Our iterative approach (Iter.) and combined iterative+parallel strategy (Iter.+Par.) consistently outperform traditional parallel-sampling baselines, with gains most pronounced on complex compositional tasks (ConceptMix k=4--7) and precise spatial and numeric reasoning (T2I-CompBench spatial, 3D spatial, and numeracy categories).}
\label{tab:combined_results}
\end{table*}

We conduct experiments with three state-of-the-art text-to-image model families -- Qwen-Image~\cite{wu2025qwen}, Gemini 2.5 Flash Image (NanoBanana), and GPT-Image-1. Qwen-Image being the open-sourced among the three. We evaluate models across three prominent compositional generation benchmarks: ConceptMix~\cite{wu2024conceptmix}, T2I-CompBench~\cite{huang2023t2i}, and TIIF-Bench~\cite{wei2025tiif}. ConceptMix measures a model’s ability to bind multiple concept categories (objects, textures, colors, shapes, styles, relations, etc.) under increasing compositional complexity, ranging from one to seven concept combinations. T2I-CompBench evaluates open-world compositionality, including attribute binding, object–object relationships, numeracy, and multi-object reasoning. TIIF-Bench focuses on fine-grained instruction following across diverse scenarios such as 3D perspective, logical negation, precise text rendering, and 2D spatial relations. We further evaluate our method on the Visual Jenga scene decomposition benchmark which tests a model's ability to progressively remove objects from a scene in a physically plausible manner.

For all benchmarks, we follow their respective evaluation protocols and use a strong multimodal language model (MMLM)  (Gemini-2.5-Pro or GPT-4o depending on dataset original specification) to assess prompt–image consistency. For ConceptMix and TIIF-Bench, question–answer prompts are provided to the evaluator MMLM along with the generated image which outputs binary/yes no answers. For T2I-CompBench, we adopt their MMLM-based scoring protocol (using GPT-4V), which outputs a continuous alignment score between 1 and 100. Importantly, our in-the-loop critic and verifier is a \textit{weaker MMLM} different from the final benchmark evaluator. For primary experiments, we use Gemini-2.5-Flash as the in-loop verifier and critic. Further implementation details and baselines are provided in appendix. 

\subsection{Compositional Image Generation}
We first evaluate on ConceptMix and T2I-CompBench. For each model (Qwen-Image, Gemini, GPT-Image), we run two variants of our method -- fully iterative (\textit{Iter}) and iterative+parallel (\textit{Iter-Par}) -- under a matched inference-time compute budget $B$. For Qwen-Image, we set $B{=}16$ for Conceptmix and $B{=}8$ for T2I-Bench (given reduced prompt complexity). Accordingly, for \textit{Iter-Par}, we consider 2 parallel steps and $B/2$ iterative steps. As detailed in Section~\ref{sec:iter-par-tradeoff}, we find this configuration to be an optimal trade-off under a fixed compute budget. As a strong budget-matched \textit{parallel-only} baseline (\textit{Parallel}), we generate $B$ images in parallel with different random seeds. We use the same VLM (Gemini-2.5-Flash) as in-loop verifier in both iterative and parallel steps to select the best image (from final set of images). This image is then passed to the benchmark-specific evaluator. Note, for GPT-Image and Gemini, we set $B{=}12$ (for Conceptmix) and $B{=}8$ (for T2I), and evaluate on a randomly sampled reduced subset of prompts for each category due to their higher inference cost and closed-source nature.

\begin{figure}[t]
    \centering
    \includegraphics[width=0.9\columnwidth]{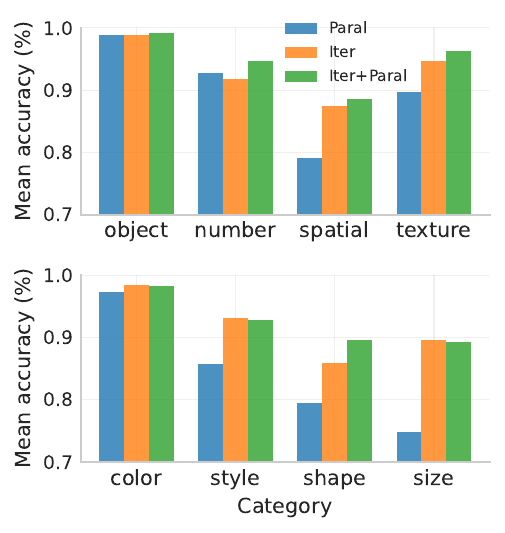}
    \caption{Per-category level improvement for ConceptMix  with Qwen-Image. As can be seen the largest improvement for iterative refinement comes from Spatial, Size, Style and Shape categories.}
    \label{fig:concept-solve-rate}
\end{figure}

\textbf{ConceptMix}: As shown in Table~\ref{tab:combined_results}, both \textit{Iter} and \textit{Iter-Par} consistently outperform the parallel-only baseline, with the largest gains on complex compositions (ConceptMix \textit{k=4--7}). For Qwen-Image, we observe improvements of 18.8\%, 21.6\%, and 16.9\% at binding complexities $k=5, 6, 7$, respectively. We also see significant gains for Nano-Banana and GPT-Image, with improvements of 8.3\% and 10.6\% at $k=7$, respectively. Notably, improvements are also present for smaller binding complexities such as $k=1, 2, 3$, indicating effectiveness even at simpler compositions. Further, in Fig.~\ref{fig:concept-solve-rate}, we show mean accuracy across ConceptMix categories for Qwen-Image; the largest improvements are observed in Spatial, Style, Shape and Size categories. In comparison, Object and Color categories do not show strong improvements. This is possibly as the model has strong capabilities for these categories and does not perform poorly for them even at high concept bindings.

\begin{table*}[t]
    \centering
    \footnotesize
    \setlength{\tabcolsep}{4pt}
    
    \begin{tabular}{l !{\vrule width 0.4pt} c !{\vrule width 0.4pt} c c c c !{\vrule width 0.4pt} c c c c c c !{\vrule width 0.4pt} c}
    \toprule
    \textbf{Model} & \textbf{Overall} &
    \multicolumn{4}{c!{\vrule width 0.4pt}}{\textbf{Basic Following}} &
    \multicolumn{6}{c!{\vrule width 0.4pt}}{\textbf{Advanced Following}} &
    \textbf{Designer} \\
    \cmidrule(lr){3-6} \cmidrule(lr){7-12} \cmidrule(lr){13-13}
    & & Avg & Attribute & Relation & Reasoning &
    Avg & Attr+Rel & Attr+Reas & Rel+Reas & Style & Text & Real World \\
    \midrule
    FLUX.1 [dev]~\cite{blackforest2024flux} & 71.1 & 83.1 & 87.1 & 87.3 & 75.0 & 65.8 & 67.1 & 73.8 & 69.1 & 66.7 & 43.8 & 70.7 \\

    SD 3~\cite{esser2024scaling} & 67.5 & 78.3 & 83.3 & 82.1 & 71.1 & 61.5 & 61.1 & 68.8 & 51.0 & 66.7 & 59.8 & 63.2 \\
    Janus-Pro-7B~\cite{chen2025januspro} & 66.5 & 79.3 & 79.3 & 78.3 & 80.3 & 59.7 & 66.1 & 70.5 & 67.2 & 60.0 & 28.8 & 65.8 \\
    MidJourney v7~\cite{midjourney2025v7} & 68.7 & 77.4 & 77.6 & 82.1 & 72.6 & 64.7 & 67.2 & 81.2 & 60.7 & 83.3 & 24.8 & 68.8 \\

    Seedream 3.0~\cite{gao2025seedream} & 86.0 & 87.1 & 90.5 & \textbf{89.9} & 80.9 & 79.2 & 79.8 & 77.2 & 75.6 & \textbf{100.0} & 97.2 & 83.2 \\
    Qwen-Parallel~\cite{wu2025qwen} & 85.2 & 85.2 & 89.7 & 88.3 & 77.7 & 80.6 & \textbf{81.9} & 79.6 & 77.8 & 89.7 & 93.7 & 90.4 \\
    \midrule
    \textbf{Qwen-Iter}     & 85.4 & 85.0 & \textbf{92.0} & 80.5 & 82.3 & 81.3 & 80.8 & 80.1 & 80.2 & 86.2 & 97.6 & 88.4 \\
    \textbf{Qwen-Iter+Par} & \textbf{87.4} & \textbf{88.1} & 90.5 & 88.1 & \textbf{85.4} & \textbf{81.5} & 81.4 & \textbf{82.0} & \textbf{80.5} & 90.0 & \textbf{97.7} & \textbf{92.0} \\
    \bottomrule
    \end{tabular}
    \caption{Performance comparison across prominent open source text-to-image models on TIFF~\cite{wei2025tiif} benchmark (short descriptions only; full long description results in suppl.). Qwen-Iter+Par achieves state-of-art and is especially beneficial on Basic Reasoning scenario as well as Attr+Reas, Rel+Reas and Text-writing categories. Compute-matched Qwen-Parallel has overall poorer performance to iterative variants. }
    \label{tab:tiff_model_comparison_short}
\end{table*}

\begin{figure}[t]
\centering
\includegraphics[width=0.98\linewidth]{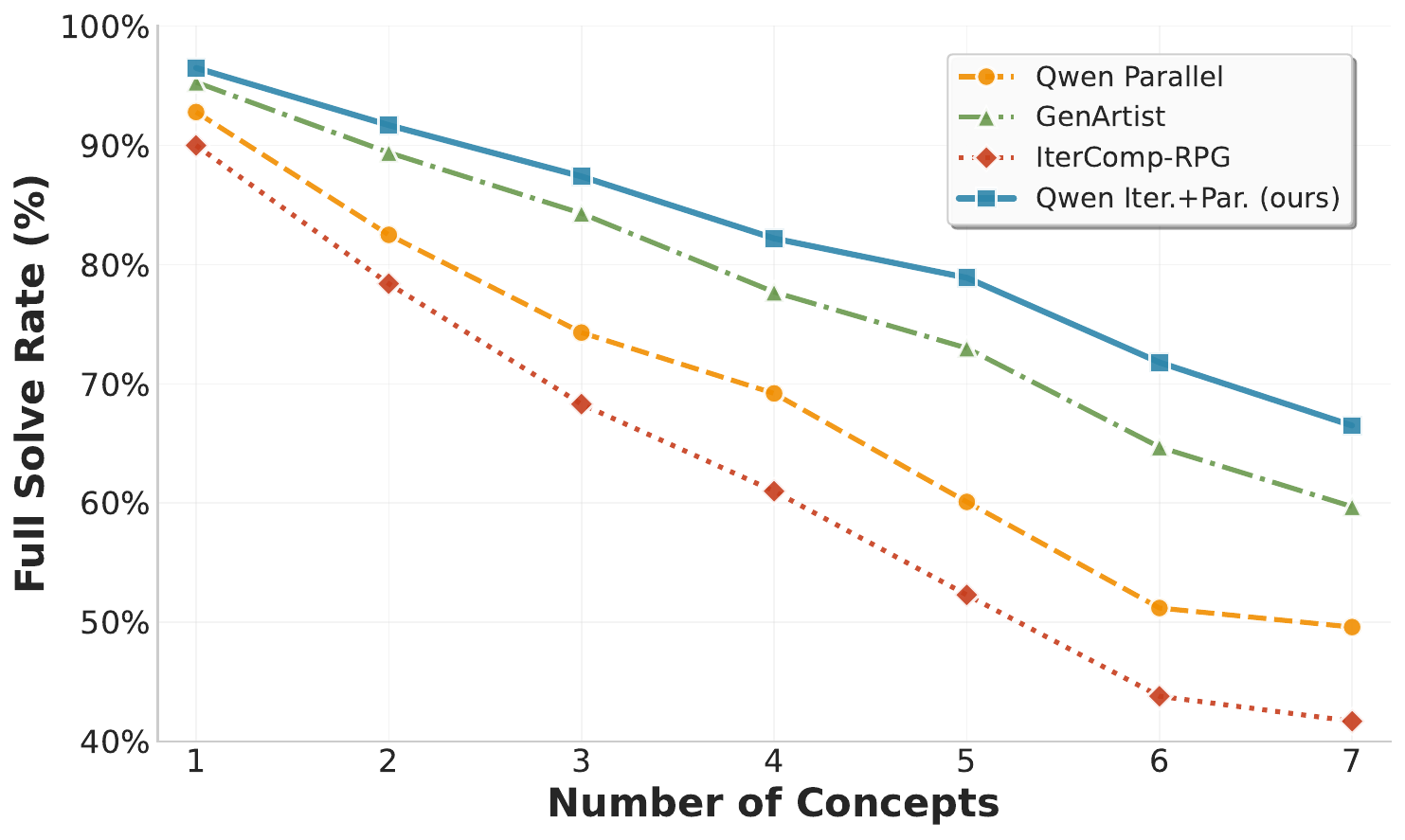}
\caption{Comparison against existing test-time scaling methods. As can be seen methods such as GenArtist \cite{wang2024genartist} and RPG \cite{yang2024mastering} are difficult to scale due to their reliance on tools or regional priors.}
\label{fig:method_comparison}
\end{figure}

\textbf{T2I-CompBench}: As shown in Table~\ref{tab:combined_results}, we observe strong gains across multiple categories of T2I-CompBench. For Qwen-Image, improvements of 7.1\%, 13.8\%, and 6.3\% are achieved on Spatial, 3D-spatial, and Numeracy, respectively, suggesting that iterative refinement particularly helps in generation of images that entail precise spatial and numeric reasoning. We observe significant gains for Nano-Banana and GPT-Image as well, with Nano-Banana improving by 6.7\%, 7.9\%, and 9.8\% on the same set of categories respectively. However, on categories such as Color, Non-Spatial and Texture we do not find strong benefits. This could be as the model is already strong in these categories and thus iterative refinement may not be required.

\textbf{Comparison with Compositional Methods.} We also compare the application of our iterative refinement method with prominent compositional generation frameworks such as IterComp~\cite{zhang2024itercomp}, RPG~\cite{yang2024mastering}, and GenArtist~\cite{wang2024genartist}. As shown in Fig.~\ref{fig:method_comparison}, while most methods have similar performance at initial binding complexities, our method shows stronger gains as the number of concepts increase. This is potentially due to our method's usage of a simple general VLM-critic and editor in loop to iteratively critique and refine the image without usage of  task-specific tools such as object detectors, super resolution generation, layout planners, etc that may introduce intermediate artifacts and compound errors over longer concept lengths. For GenArtist, we maintain the same base model Qwen-Image, while for RPG, we use the recent IterComp implementation~\cite{zhang2024itercomp} which uses stable diffusion as the base model. Additional details and baseline discussions are provided in the supplemental.

\begin{figure*}[t]
    \centering
    \includegraphics[width=0.98\linewidth, trim=500 300 500 80,clip]{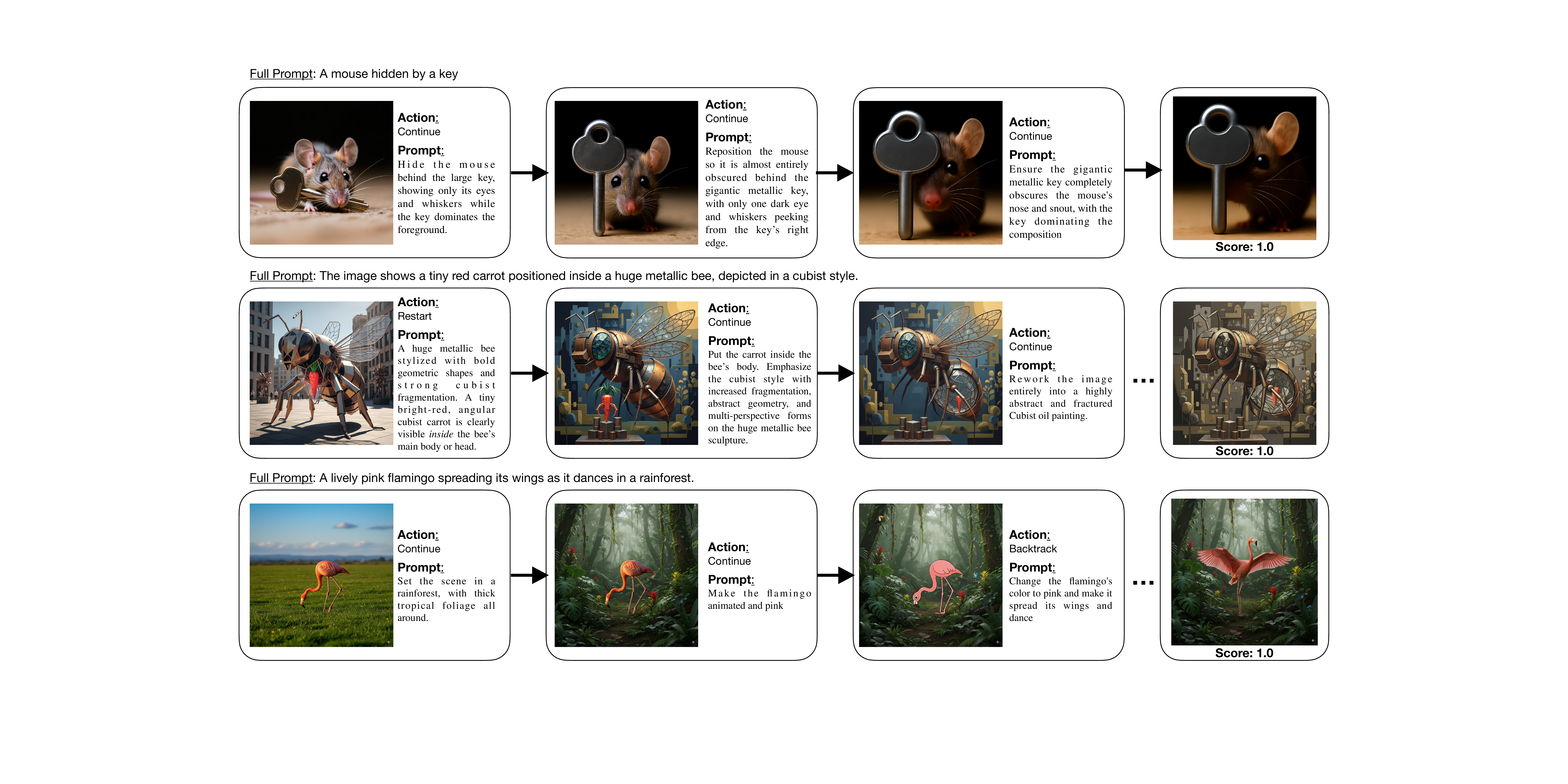}
    \caption{Each row shows a different prompt and the corresponding sequence of critic-guided refinement steps. The critic issues actions (Continue, Restart, Backtrack) and targeted sub-prompts, allowing the generator to progressively correct errors—e.g., hiding the mouse behind the key, placing the carrot inside the metallic bee in a cubist style, and adjusting the flamingo’s pose. The final images satisfy all compositional constraints with high fidelity. See appendix for more examples and \textbf{failure cases} (figs. \ref{fig:failure1} and \ref{fig:failure2}).}
    \label{fig:qualitative}
\end{figure*}

\begin{figure}[t]
\centering
\includegraphics[width=0.99\linewidth]{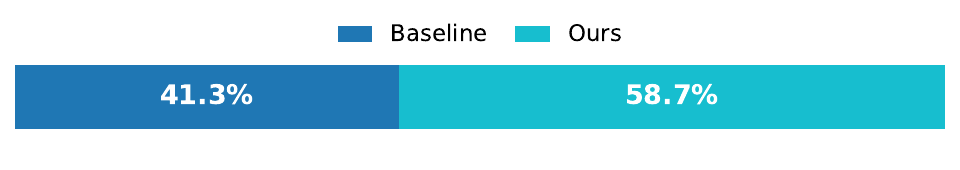}
\vspace{-1em}
\caption{Human preference rate. As can be seen our method is preferred over the parallel only baseline by the human evaluators.}
\label{fig:preference}
\end{figure}

\textbf{TIIF-Bench:} In Table~\ref{tab:tiff_model_comparison_short}, we report results on the TIIF-Bench benchmark which evaluates basic and advanced instruction-following capabilities of image generation models. We set Qwen-Iter+Par achieves state-of-the-art results among open-source methods, including a 5.0\% improvement over Qwen-Parallel on basic reasoning prompts, 2.7\% on advanced Relation+Reasoning, and 4.0\% on text rendering. These results underscore the generality of our method and its applicability across diverse compositional text-to-image scenarios requiring varied reasoning skills.

\subsection{Qualitative analysis and human evaluation} 
\label{sec:qual}
Here, we qualitatively analyze intermediate generations alongside the critic’s refinement prompts and actions. In Fig.~\ref{fig:qualitative}, we show three examples covering different refinement patterns. In the first example (prompt: \textit{“A mouse hidden by a key”}), the initial image incorrectly depicts the mouse holding the key. The critic selects Continue and proposes a refinement prompt to hide the mouse behind the large key so only whiskers and eyes are visible. The next edit still leaves the mouse only partially hidden, so the critic again selects Continue and emphasizes that the mouse should be almost entirely obscured behind the metal key. The subsequent image improves but remains unconvincing, prompting another Continue action with a refinement to ensure the key dominates the composition and fully obscures the mouse. The final image convincingly depicts a mouse hidden by a key.

In the second example (prompt: \textit{“a tiny red carrot positioned inside a huge metallic bee, depicted in a cubist style”}), the initial image lacks the cubist style. The critic chooses Restart (Fresh Start), yielding a more cubist rendering but with the carrot outside the bee. The critic then selects Continue and refines the prompt to place the carrot inside the bee while reinforcing the cubist style. The next image improves on both criteria, and a final Continue instructs a highly abstract, cubist oil-painting treatment. The resulting image convincingly shows a tiny red carrot inside a metallic bee in a cubist style.

Finally, in the third example (prompt: “a lively pink flamingo spreading its wings as it dances in a rainforest”), the initial image incorrectly shows a flamingo grazing in a grass field. The critic selects Continue and refines the prompt to set the scene in a rainforest; the next image places the flamingo there. The critic then proposes to make the flamingo more animated, but this yields an overly “animated” stylization rather than a lively pose. The critic Backtracks to the previous image and refines the prompt to make the flamingo lively—explicitly asking it to spread its wings and dance in the rainforest. The resulting image shows a wing-spread, dancing flamingo in a rainforest, closely matching the prompt.

\textbf{Human Evaluation:} We conducted a user study on 150 randomly sampled prompts from ConceptMix and T2I-CompBench. For each prompt, three raters answered a set of questions about the generated image, related to the correctness of each concept, attribute and relation mentioned in the prompt. We also collected preference judgments by presenting two images side-by-side along with the text prompt and asking raters to select their preferred image. A sample UI and additional details are provided in the supplemental. As shown in Fig.~\ref{fig:preference}, our method is preferred 58.7\% of the time versus 41.3\% for the parallel-only baseline. Inter-annotator agreement among humans is 85.3\%. The average agreement between humans and the language model for the same set of images is 83.4\%, indicating that the language model based evaluator (MMLM) is sufficiently reliable.

\begin{figure*}[t]
    \centering

    \includegraphics[width=0.99\linewidth]{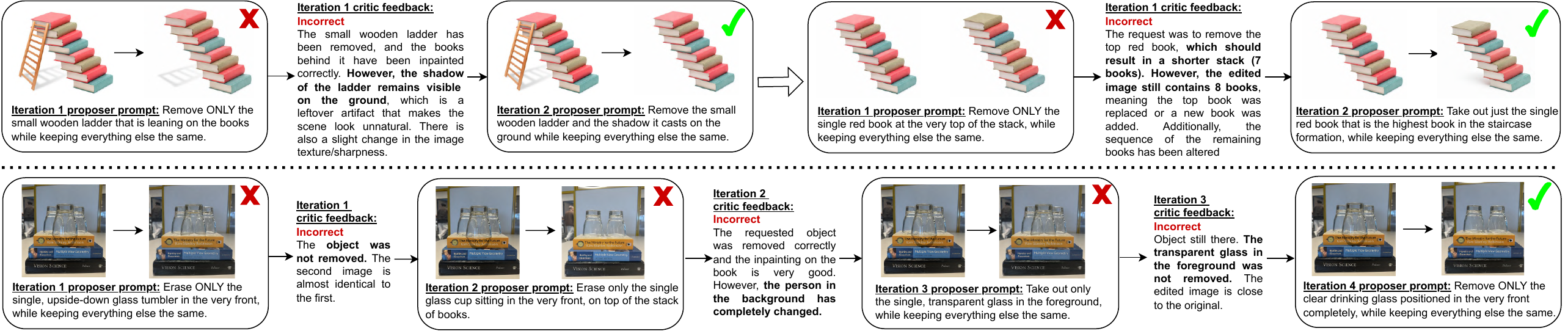}
    \caption{Qualitative analysis on Visual Jenga \cite{bhattad2025visual} scene decomposition task. The \textbf{top row} shows how the VLM critic is able to identify errors (such as shadow residual artifact in first case or incorrect book removal in second case) and issue corrective feedback that the proposer incorporates for next step instruction. The \textbf{bottom row} shows an example where three rounds of feedback are required to generate correct removal with critic able to identify even subtle changes such as change of person appearance in background (see iteration 2).} 
    \label{fig:qualitative_vis_jenga}
\end{figure*}

\subsection{Visual Jenga Scene Decomposition}
\label{sec:visual-jenga}
Here, we extend our iterative refinement framework to the recent Visual Jenga \cite{bhattad2025visual} full scene decomposition task, where the goal is to progressively remove objects from a cluttered scene in a \emph{physically plausible} sequence while maintaining correctness at every intermediate generation. Starting from an initial scene image, the system selects the next object to remove, produces a \emph{removal phrase} (e.g., ``remove the red mug from the table''), and generates the next scene representation with that object removed.

\textbf{Extending our iterative refinement setup.} At each step, given the current scene, a VLM proposer suggests the next object and corresponding removal phrase, which is passed to the editor to generate the next scene. We then feed \emph{both} the previous and next images to a VLM critic that checks (1) the specified object was removed correctly, and (2) no other violations occurred (hallucinations, artifacts, identity drift, or physically implausible changes). If a violation is detected, the critic returns structured feedback that is fed back into the proposer to generate a more precise removal phrase or to choose an alternate next object, and the step is retried. This continues until the step is verified by the critic or the per-step compute budget is exhausted (in which case the highest scoring candidate is selected). We report results using the recent GPT-Image-1.5.

We compare against a budget-matched \textit{parallel sampling} baseline that generates 4 candidates per removal step and uses the same VLM critic as verifier to select the best candidate. We evaluate on the full decomposition subset consisting of 56 unique scenes, and report the \emph{full solve rate}: a scene is counted as solved only if the method completes the entire decomposition sequence with all intermediate steps satisfactory to human evaluation. As shown in Table~\ref{tab:visual_jenga}, applying iterative feedback through VLM critic feedback improves full decomposition solve rate from 64.29\% (with parallel sampling) to 76.79\%.

In Fig.~\ref{fig:qualitative_vis_jenga}, we illustrate how the VLM critic identifies errors in candidate removals and how the proposer incorporates this feedback to correct the removal phrases. In the top row first case, removing only the small wooden ladder leaves behind its shadow; the VLM critic identifies this in its feedback, and in the next iteration, the proposer outputs prompt to remove both \textit{the ladder and the shadow it casts}, producing a correct intermediate scene generation. The second case in the top row similarly shows the critic detecting incorrect removal of the top red book (with the count of books unchanged) and providing precise corrective feedback. The bottom case shows how three rounds of feedback progressively refine the prompt until the \textit{frontmost glass} is correctly deemed by critic to be successfully removed.

Overall, these results further underscore the benefit of iterative refinement in providing corrective feedback to improve intermediate generations, which is not captured in compute-matched parallel sampling thereby leading to lower performance in scene decomposition as well.

\begin{table}[t]
    \centering
    \footnotesize
    \setlength{\tabcolsep}{8pt}
    \begin{tabular}{l|cc}
        \toprule
        & \textbf{GPT-Parallel} & \textbf{GPT-Iter (ours)} \\
        \midrule

        Full solve rate (\%) ($\uparrow$) & 64.29 & 76.79 \\
        \bottomrule
    \end{tabular}
    \caption{Results on Visual Jenga full scene decomposition. Iterative refinement significantly improves full solve rate over compute-matched parallel sampling.}
    \label{tab:visual_jenga}
\end{table}

\begin{table}[t]
    \centering
    \footnotesize
    \setlength{\tabcolsep}{5pt}
    \begin{tabular}{c c c c c}
    \toprule
    $I$ & $P$ & $I \times P$ & CMix(k$\{5,6,7\}$) & T2I-Avg\\
    \midrule
    1  & 1  & 1  & 32.3 & 79.8 \\
    \midrule
    1  & 2  & 2  & 35.6 & 82.3 \\
    2  & 1  & 2  & \textbf{37.2} & \textbf{82.6} \\
    \midrule
    1  & 4  & 4  & 41.1 & 84.9 \\
    2  & 2  & 4  & 41.3 & 84.7 \\
    4  & 1  & 4  & \textbf{48.4} & 86.4 \\
    \midrule
    1  & 8  & 8  & 44.8 & 86.5 \\
    2  & 4  & 8  & 43.9 & 87.4 \\
    4  & 2  & 8  & 57.6 & \textbf{90.2} \\
    8  & 1  & 8  & \textbf{60.0} & 89.9 \\
    \midrule
    1  & 16 & 16 & 52.1 & 87.9 \\
    2  & 8  & 16 & 53.4 & 89.0 \\
    4  & 4  & 16 & 66.3 & 91.7 \\
    8  & 2  & 16 & \textbf{69.6} & \textbf{92.6} \\
    16 & 1  & 16 & 69.2 & 92.1 \\
    \bottomrule
    \end{tabular}
    \caption{\textbf{Average accuracy} of configurations sorted by total compute; $I$ = iterative steps, $P$ = parallel steps. A higher proportion of iterative compute w.r.t. to parallel compute consistently leads to better results across different computation budgets.}
    \label{tab:iter-par-acc}
\end{table}

\subsection{Ablations}
\label{sec:iter-par-tradeoff}

\textbf{Trade-offs between iterative and parallel compute.} We analyze the trade-offs between iterative and parallel computation under different inference-time compute budgets. Given a total budget $B$, we study the allocation between iterative steps $I$ and parallel steps $P$ (where $I \times P = B$). As shown in Table~\ref{tab:iter-par-acc}, we evaluate $(I,P)$ configurations for $B \in \{1,2,4,8,16\}$. Due to the high computational cost entailed, we conduct this analysis on the open-source Qwen-Image model, using a randomly sampled reduced subset of prompts from ConceptMix (binding lengths $k \in \{5,6,7\}$) and T2I-CompBench (complex, 3D spatial, numeracy, spatial, color, texture). Across larger budgets ($B \ge 4$), iterative allocation yields higher accuracy than parallel. For $B{=}4$, purely iterative ($I{=}4, P{=}1$) achieves 48.4\% on ConceptMix and 86.4\% on T2I-CompBench, compared to 41.1\% and 84.9\% for purely parallel ($I{=}1, P{=}4$). Similarly, fully-iterative ($I{=}B, P{=}1$) outperforms fully-parallel ($I{=}1, P{=}B$) on ConceptMix by 15.2\% and 17.1\% at $B{=}8$ and $B{=}16$, respectively.

\begin{figure}[t]

    \centering
    \includegraphics[width=0.9\columnwidth]{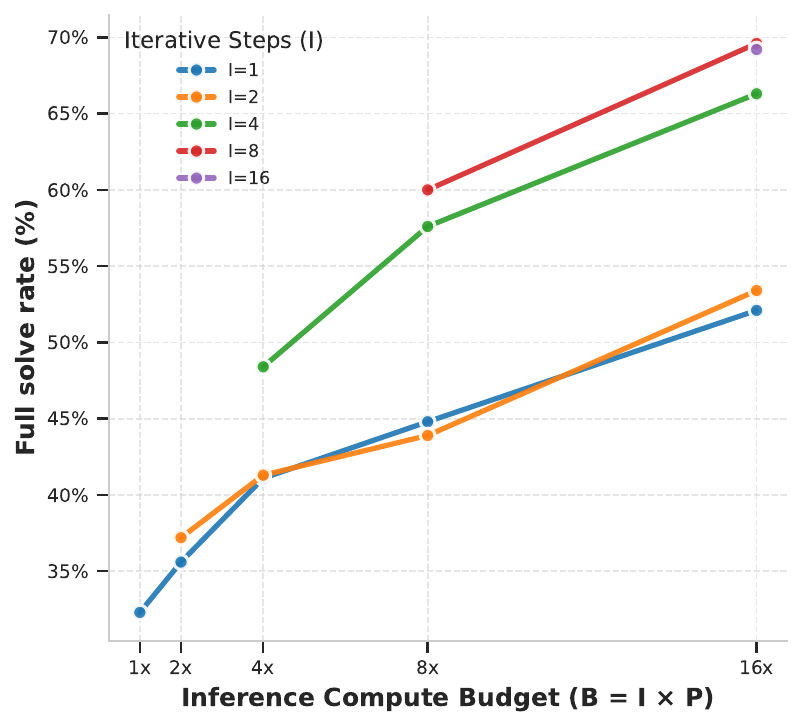}
  \caption{Comparison of iterative and parallel compute allocations.
Given a budget of $B=16$, mixed allocations of 8 iterative with 2 parallel generally outperform purely parallel or purely iterative strategies.}

    \label{fig:iter-par-comp}
\end{figure}

At $B{=}16$, the best allocation is $I{=}8, P{=}2$, achieving 69.6\% on ConceptMix and 92.6\% on T2I-CompBench, compared to $I{=}16, P{=}1$ (69.2\% and 92.1\%). This indicates that, at higher budgets, a mixed strategy -- primarily iterative refinement with a small amount of parallel sampling -- can outperform purely iterative or purely parallel approaches. We believe this could be due to diminishing returns beyond a certain number of iterative refinements. Instead of doing unnecessary refinement steps, allocating a portion of compute to parallel candidates improves exploration and prevents over-refinement. Note that even within mixed allocations, skewing the budget toward iteration rather than parallelism performs best (Table~\ref{tab:iter-par-acc}). As shown in Fig.~\ref{fig:iter-par-comp}, settings with higher $I$ (green and red lines denoting $I{=}4$ and $I{=}8$, respectively; purple dot denoting $I{=}16$) consistently achieve higher solve rates than settings with lower $I$ (blue and orange lines denoting $I{=}1$ and $I{=}2$) at the same budget $B$. Overall, these results indicate that the combination of iterative refinement and parallel sampling is an optimal setting for compositional image generation.

\textbf{Choice of VLM critic model and action space} Here, we analyze the impact of the backbone VLM critic model and the action space of the critic model.  As shown in table.~\ref{tab:critic-vlm-ablation}, we analyzed usage of Gemini-Pro, GPT-5 and Qwen3-VL-32B-Instruct models as the critic models for our method on a subset of randomly sampled Conceptmix prompts. As shown, with Gemini-Pro as the critic model, we achieve $\sim$  4\% performance improvement than our default used Gemini-2.5-Flash model. However, using the small open-source Qwen3-VL-32B shows $\sim$ 3\% performance degradation indicating that recent improvements in the foundation model capabilities are critical to our improvement. In Table \ref{tab:action-space-ablation}, we analyze the impact of the action space of the critic model. As shown, we achieve the best accuracy with the full action space, with both actions of \textit{`Fresh Start'} and \textit{`Backtrack'} being beneficial.

\begin{table}[t]
    \centering
    \footnotesize
    \setlength{\tabcolsep}{3pt}
    \begin{tabular}{l c}
    \toprule
    \textbf{Critic VLM} & \textbf{Solve rate (\%).} \\
    \midrule
    Gemini-2.5-Flash & 69.7 \\ 
    Gemini-Pro & 74.0 \\ 
    GPT-5 & 72.3 \\ 
    Qwen3-VL-32B & 66.3 \\
    \bottomrule
    \end{tabular}
    \caption{Impact of choice of critic VLM on performance.}
    \label{tab:critic-vlm-ablation}
\end{table}

\begin{table}[t]
    \centering
    \footnotesize
    \setlength{\tabcolsep}{3pt}
    \begin{tabular}{l c}
    \toprule
    \textbf{Action Space} & \textbf{Solve rate (\%).} \\
    \midrule
    Full action space & 69.7 \\
    w/o Backtrack & 68.0 \\
    w/o Fresh Start & 67.7 \\
    w/o Backtrack \& Fresh Start & 67.3\\
    \bottomrule
    \end{tabular}
    \caption{Impact of action space components on performance.}
    \label{tab:action-space-ablation}
\end{table}

\vspace{-0.4em}
\section{Conclusion}
We introduced iterative refinement as a simple but broadly applicable  inference-time strategy to improve compositional image generation capabilities of text-to-image (T2I) models. Our approach combines an image generation and editing model with a vision-language model (VLM) critic in the loop that progressively enables refinement of generated outputs. We show our method achieves strong performance benefits over traditional inference-time scaling methods such as parallel sampling on prominent T2I models including Nano-Banana, Qwen-Image and GPT-One. Our framework achieves state-of-art performances across compositional image generation benchmarks including Conceptmix, T2I-CompBench and TIFF as well as the Visual Jenga scene decomposition task. We further perform qualitative analysis to illustrate how our method works, besides human evaluation to concretely verify our method beyond benchmarks. We also conduct ablations studying tradeoffs between iterative and parallel compute allocation, and the impact of choice of VLM critic model and its action space.

\newpage
{
    \small
    \bibliographystyle{ieeenat_fullname}
    \bibliography{main}
}

\clearpage
\maketitlesupplarxiv

\section{Further qualitative examples.} We visualize further samples in the attached interactive HTML webpage for different prompt types showcasing our method (Iterative Refinement) compared to the baseline (best chosen sample out of 16 parallel samples) for 3 model families -- Qwen-Image, GPT and NanoBanana. 

\textbf{Limitations and selected failure modes}
Our qualitative analysis reveals two primary failure modes:

\textbf{(i) Incorrect VLM reasoning:} When the VLM critic or verifier produces faulty reasoning, it generates an incorrect verification signal. This can cause genuine errors in the generated image to go undetected, or conversely, lead to unnecessary refinements of correct images.

\textbf{(ii) Inability of editor to make prompted changes:} We also observed cases where the editor was unable to make the desired changes to the image, even though the prompt was clear. This is likely due to the complexity of the image and the limitations of the image editing model. The prompt ``The image features a heart-shaped giraffe, a tiny pink screwdriver, and a huge robot. The screwdriver is positioned at the bottom of the robot, touching it.'' is an example of a prompt where the editor was unable to make the desired changes.

\begin{figure}
    
\begin{center}
  \includegraphics[width=0.99\linewidth]{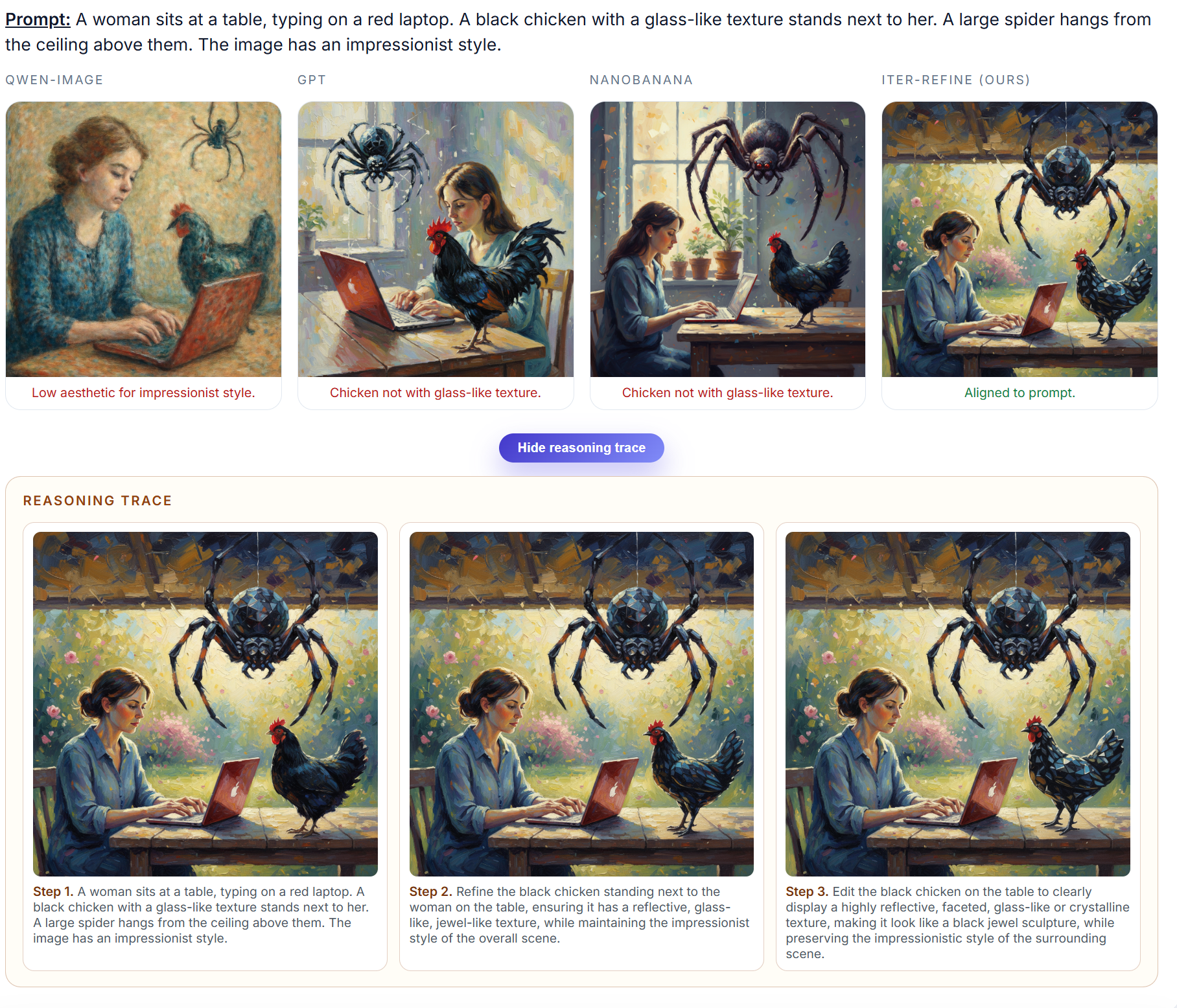}
\end{center}
\caption{Example of model correcting `glass-shaped' chicken over steps (see full example in visualization html page).}
\label{fig:failure1}
\end{figure}

\begin{figure}
    
\begin{center}
  \includegraphics[width=0.99\linewidth]{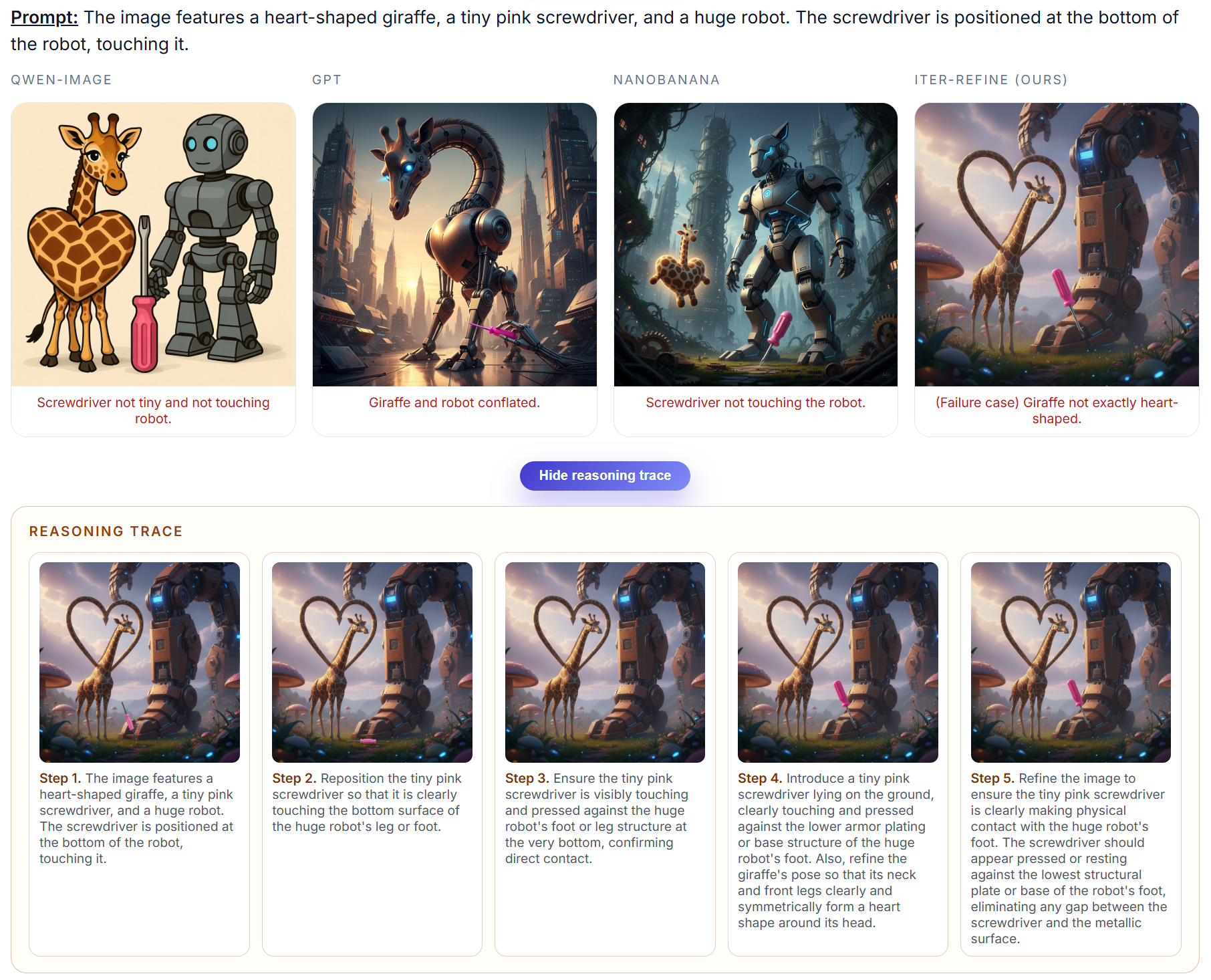}
\end{center}
\caption{Example of a failure case due to verifier not detecting giraffe is not heart shaped (see full example in visualization html page).}
\label{fig:failure1}
\end{figure}

\begin{figure}[h]
\begin{center}
  \includegraphics[width=0.99\linewidth]{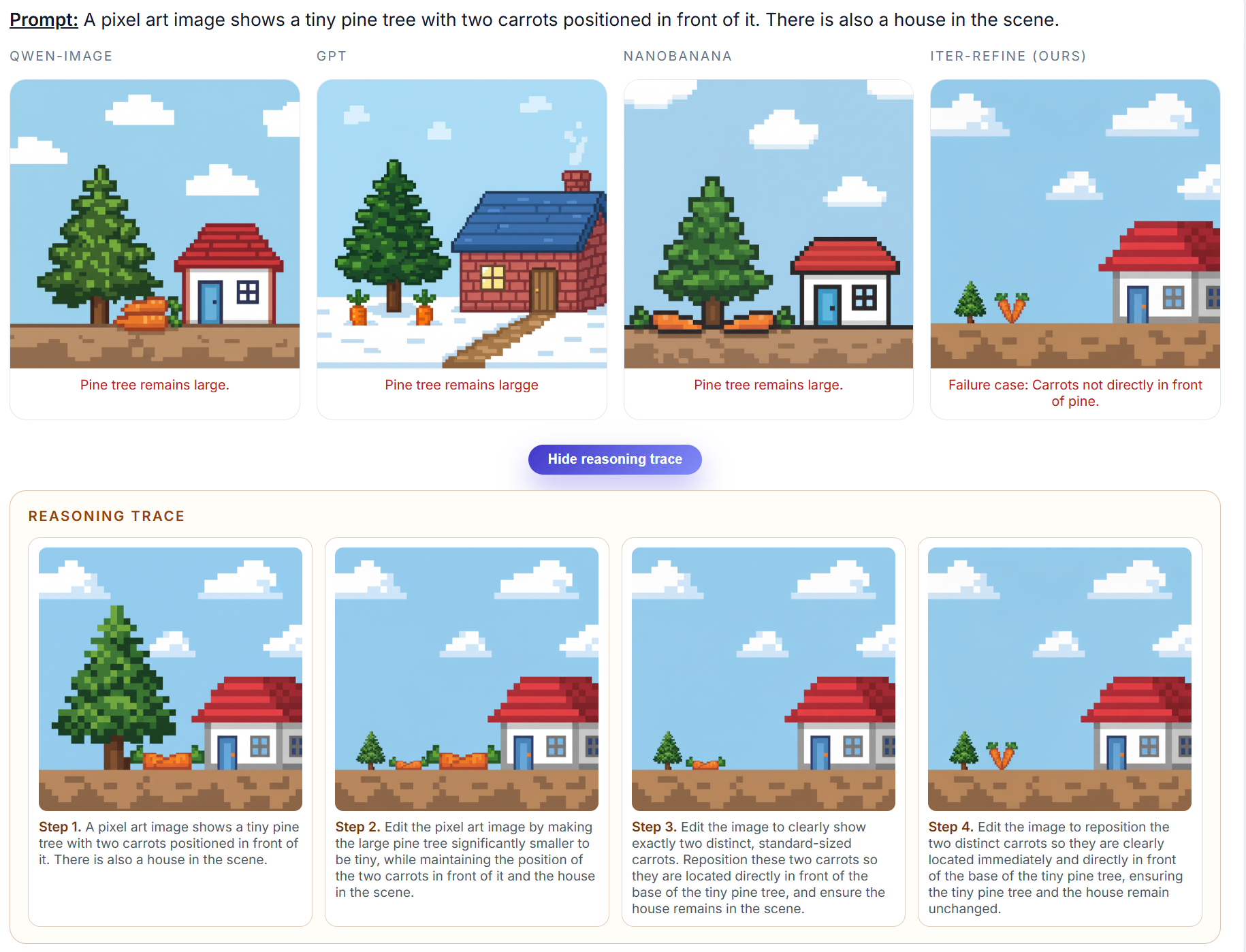}
\end{center}
\caption{Example of a failure case where editor is not able to add carrot in front of pine tree.}
\label{fig:failure2}
\end{figure}

\section{Further experiment details}
We provide further experiment details here regarding models used and how they were accessed. We will open source our codebase for reproducability.

\textbf{Image Generation Models:}
\begin{itemize}
    \item \textbf{Qwen-Image} and \textbf{Qwen-Image-Edit}: Weights obtained from the Hugging Face model hub.
    \item \textbf{GPT-Image-One}: Run on 'auto' inference setting (for cost reasons) with official closed-source OpenAI API using model id: \texttt{gpt-image-one}.
    \item \textbf{NanoBanana}: Run on default inference setting with Google GenAI API using model id: \texttt{gemini-2.5-flash-image}.
\end{itemize}

\textbf{Verifier and Critic Models:}
\begin{itemize}
    \item \textbf{Gemini-2.5-Flash-Lite}: Used for our primary experiments as the in-loop critic and verifier with Google GenAI API using model id: \texttt{gemini-2.5-flash-lite}.
    \item \textbf{Gemini-2.5-Pro}: Used for ablation studies and as final evaluation verifier for ConceptMix.
    \item \textbf{GPT-4o}: Used for T2I-CompBench and TIIF-Bench as official VLM evaluator (following benchmark protocol).
    \item \textbf{GPT-5-Mini}: Used for ablation on 'auto' thinking complexity (for cost reasons) with official closed-source OpenAI API using model id: \texttt{gpt-5}.
    \item \textbf{Qwen3-VL-32B-Instruct}: Used for ablation studies with weights obtained from official Hugging Face model hub.
\end{itemize}

We provide system and user prompts provided to the VLM critic at each turn below.

\begin{tcolorbox}[
  title={User Prompt (Provided to critic after each iteration)},
  colback=gray!5,
  colframe=gray!70
]
    \textbf{Full complex prompt:}  
    In an abstract ink style image, a corgi stands near a large tree. Nearby, there are three tiny hills with a metallic texture.
    
    \medskip
    \textbf{Your previous step prompts were:}
    \begin{itemize}
      \item Step 1: In an abstract ink style image, a corgi stands near a large tree. Nearby, there are three tiny hills with a metallic texture.
      \item Step 2: Change the texture of the three hills in the foreground to be shiny and metallic.
    \end{itemize}
    
    \medskip
    \textbf{The most recently generated image had the following verifier scores:}
    \begin{itemize}
      \item Does the image contain corgi?: 1
      \item Does the image contain hills?: 1
      \item Does the hill have a metallic texture?: 1
      \item Is the style of the image abstract?: 0
      \item Is the style of the image ink?: 1
      \item Is the hill tiny in size?: 1
      \item Is the number of hills exactly 3?: 1
      \item Cumulative mean binary score: 0.857
    \end{itemize}
    
    \medskip
    The maximum number of editing steps is \textbf{4}.  
    This is \textbf{step 3} of image editing and you will have \textbf{1 step left} to complete the task.  
    Decide the next step prompt accordingly.
\end{tcolorbox}

\begin{tcolorbox}[title={System Prompt for Critic}]
You are a helpful assistant that given a complex image generation prompt and previously generated image along with verifier scores, generates the best next step prompt for an image editing model.

The idea is to generate the image over multiple editing and refinement steps, so the next step prompt should either edit the previous image to improve it or add new elements to the image. Some suggested guidelines are:
\begin{itemize}
  \item Check if previous step worked correctly
  \item Identify any important missing element from full prompt
  \item Check if there is space for new elements to be added in the current frame. If not, then prompt model to zoom out and make space first.
  \item In case of errors, prompt model to fix them or delete the incorrect element.
\end{itemize}

You have to choose from the following actions:
\begin{enumerate}
  \item CONTINUE: Continue editing the most recently generated image to improve it with your proposed prompt.
  \item BACKTRACK: Backtrack to image before the most recently generated image, and edit that image with your proposed prompt.
  \item FRESH\_START: Start entirely from scratch with your proposed prompt due to major unfixable errors over steps.
  \item STOP: Stop the editing process due to completion of the task
\end{enumerate}

You will be provided following inputs:
\begin{itemize}
  \item The full complex prompt
  \item Your previously proposed step prompts
  \item The most recently generated image  (which is attached for your reference) along with verifier scores (sometimes verifier can be wrong for attribute counts questions)
\end{itemize}

You have to output two things:
\begin{enumerate}
  \item The action to be taken
  \item The next step prompt that will be given to the image editor or generator
\end{enumerate}

The maximum number of editing steps is 4. This is step 1 of image editing and you will have 3 steps left to complete the task. Decide the next step prompt accordingly.

\textbf{Output your response in the following format:}
\begin{verbatim}
Action: [action to be taken]
Prompt: [next step prompt]
\end{verbatim}

\end{tcolorbox}

\textbf{Further compositional generation baselines.}
We compare against prominent methods including IterComp~\cite{zhang2024itercomp}, RPG, and GenArtist~\cite{wang2024genartist}. For IterComp and RPG, we use their official codebases and checkpoints from \url{https://github.com/YangLing0818/IterComp} and \url{https://github.com/YangLing0818/RPG-DiffusionMaster}, respectively. For GenArtist, we use code from \url{https://github.com/zhenyuw16/GenArtist} and initialize the image generator and editor with Qwen-Image and Qwen-Edit for fair comparison, which we empirically found to perform better than their original Stable Diffusion \cite{meng2021sdedit} models. We adopt all other tools as specified in their original codebase.

\subsection{Human evaluation details}
\begin{figure}[t]
\begin{center}
  \includegraphics[width=0.90\linewidth]{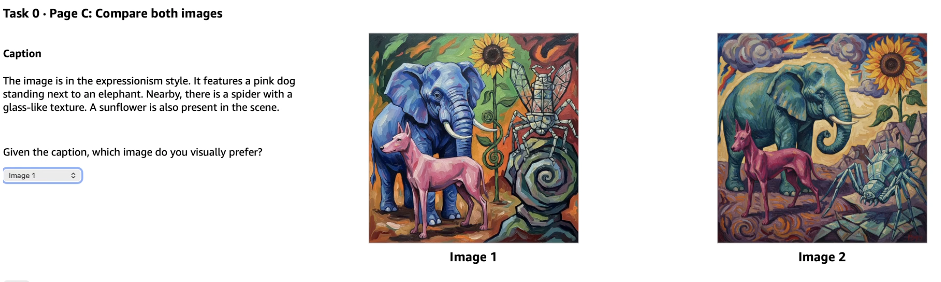}
\end{center}
\caption{User interface for the pairwise preference question showing two generated images side-by-side along with the corresponding text prompt.}
\label{fig:form_prefers}
\end{figure}

\begin{figure}[t]
\begin{center}
  \includegraphics[width=0.99\linewidth]{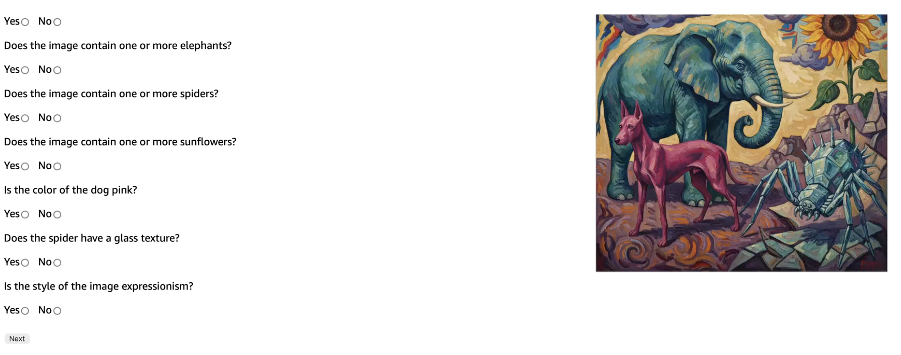}
\end{center}
\caption{User interface for the yes/no evaluation questions used to assess whether each concept, attribute, and relation in the prompt is correctly depicted in the generated image.}
\label{fig:form_yesno}
\end{figure}

To conduct our human evaluation for the images, we first selected a random subset of 150 images from ConceptMix~\cite{wu2024conceptmix}. From the $k{=}5,6,7$ groups we selected 45, 50, and 55 images, respectively. We then generated images for these prompts using the Qwen~\cite{wu2025qwen}  model with inference-time budget of 16 computational steps under two settings: (i) baseline best parallely sampled image, and (ii) our iterative method's best refined image. We defined each image's score as the number of evaluation questions the model answered "yes" to out of the total number (5, 6, or 7) of evaluation questions for that image. This produced 150 pairs of images, each with an associated score. We randomly shuffled these pairs and partitioned them into 25 blocks of 6 image-pair tasks. For each block, three participants evaluated every image using that image's ConceptMix yes/no evaluation questions. Within each pair, the images generated by our iterative method and by the parallel sampling method were shown in random order. After answering the yes/no questions for each image, participants were also asked, given the prompt, which of the two images they preferred. Examples of our form UI can be seen in Figure~\ref{fig:form_prefers} and Figure~\ref{fig:form_yesno}.
In total, we collected responses from 75 unique participants (3 participants per block), providing the data used in our human evaluation analysis.

We computed several key metrics from this human evaluation study. As seen in Table \ref{tab:human_solve_rates}, Human evaluators achieved a mean net solve rate of 91.0\% and a mean perfect solve rate of 55.7\% across all participants. For human--model agreement, images generated with 8 iterative refinement steps showed higher agreement with human evaluators (88.9\% net agreement, 30.6\% perfect agreement) compared to the highest-rated image out of 8 images generated in parallel with a single iterative step each (82.1\% net agreement, 18.7\% perfect agreement). In Table \ref{tab:human_iter_agreement}, when comparing the two generation methods directly, the net agreement for images produced with 8 iterative steps is slightly higher than for images generated in parallel with a single step (86.1\% vs.\ 84.8\% net agreement), and our method's images achieve higher perfect agreement rates as well (37.3\% vs.\ 33.3\%). This preference is most pronounced for $k{=}6$ prompts (66.7\% preference for our method) and somewhat weaker for $k{=}5$ (57.3\%) and $k{=}7$ (54.7\%) prompts.

\begin{table}[t]
    \centering
    \footnotesize
    \setlength{\tabcolsep}{4pt}

    \begin{tabular}{l c c c c}
    \toprule
    \textbf{Participant} & \textbf{Net Solve Rate} & \textbf{Images Perfect} & \textbf{Perfect Solve Rate} \\
    \midrule
    Participant 1  & 0.905 & 165/300 & 0.550 \\
    Participant 2 & 0.912 & 169/300 & 0.563 \\
    Participant 3 & 0.912 & 167/300 & 0.557 \\
    \midrule
    \textbf{Mean} & \textbf{0.910} &  & \textbf{0.557} \\
    \bottomrule
    \end{tabular}
    \caption{Human evaluator solve rates across all participants. Net solve rate represents the percentage of questions marked as `correct' by evaluator for given images, while perfect solve rate represents the percentage of images where all questions were answered correctly.}
    \label{tab:human_solve_rates}
\end{table}

\begin{table}[h]
    \centering
    \footnotesize
    \setlength{\tabcolsep}{4pt}

    \begin{tabular}{l c c c}
    \toprule
    \textbf{Image Type} & \textbf{Agreed Questions} & \textbf{Total Questions} & \textbf{Percentage} \\
    \midrule
    \multicolumn{4}{c}{\textbf{Perfect Agreements}} \\
    \midrule
    Para. & 50 & 150 & 33.3\% \\
    Iter. & 56 & 150 & 37.3\% \\
    \midrule
    \multicolumn{4}{c}{\textbf{Net Preferences by $k$ Category}} \\
    \midrule
    \textbf{$k$ Category} & \textbf{Para. } & \textbf{Iterative} & \textbf{Iter. vs Para.} \\
    \midrule
    $k{=}5$ & 42.7\% & 57.3\% & +14.6\% \\
    $k{=}6$ & 47.3\% & 64.7\% & +17.4\% \\
    $k{=}7$ & 45.3\% & 54.7\% & +9.4\% \\
    \bottomrule
    \end{tabular}
    \caption{Human-iteration agreement analysis comparing images generated with 1 iterative step vs.\ images generated with 8 iterative steps. Perfect agreements represent images where all human evaluators agreed on all questions. Net agreements represent total agreed questions across all evaluators. Preferences show the percentage of cases where each method was preferred by humans for different complexity levels ($k$), and the rightmost column reports the improvement of 8 iterative steps over 1 iterative step.}
    \label{tab:human_iter_agreement}
\end{table}

\subsection{Additional experiment specifications}
\begin{figure}[t]
\centering
\includegraphics[width=0.98\linewidth]{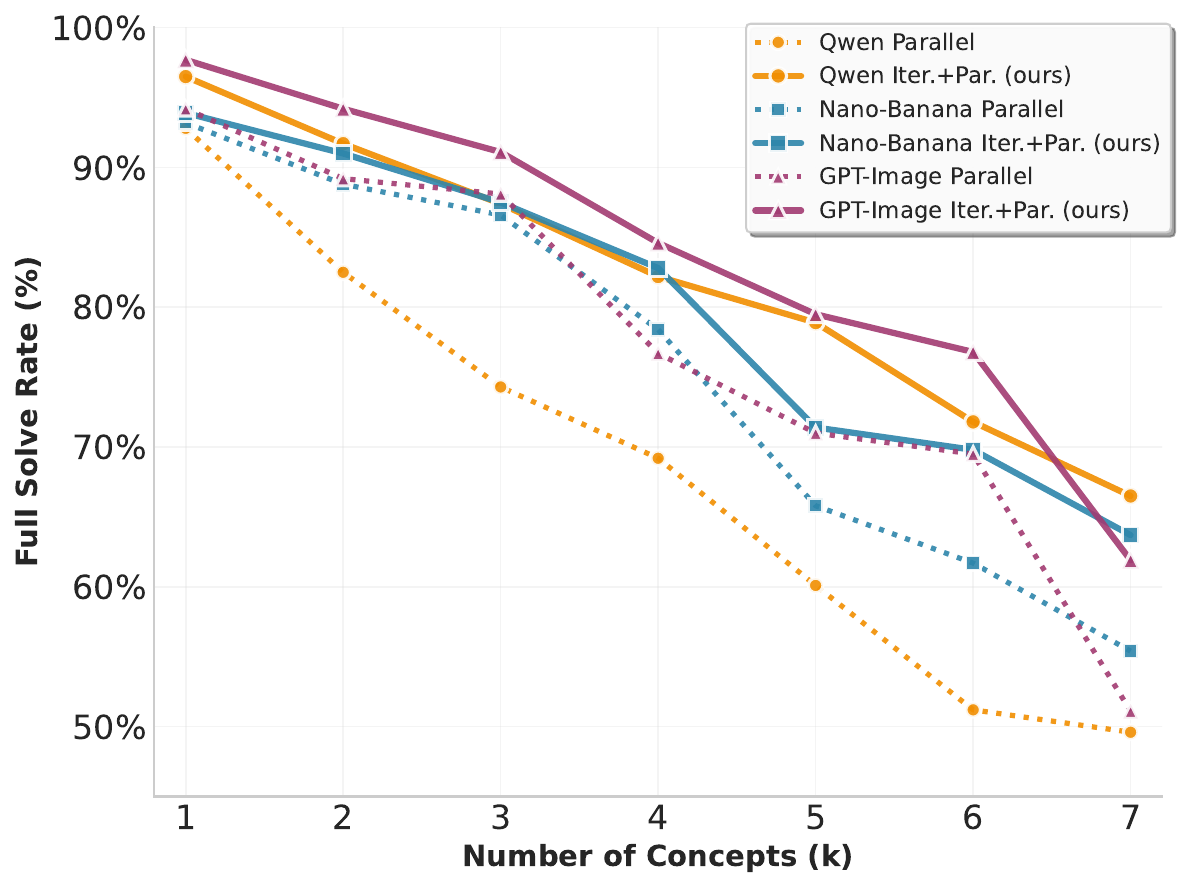}
\caption{Performance of experimented models on ConceptMix k=1 to k=7 comparison for different models. As shown, our method consistently improves over the baseline across models and prompt complexities.}
\label{fig:conceptmix_comparison}
\end{figure}

We show cumulative results over different models for ConceptMix $k=1$ to $k=7$ in Figure~\ref{fig:conceptmix_comparison}. As illustrated, our iterative refinement approach consistently outperforms the parallel-only baseline across all three models (Qwen-Image, Nano-Banana, and GPT-Image) and across all prompt complexity levels. The performance benefits are most pronounced for Qwen-Image, where we observe substantial improvements particularly at higher binding complexities ($k=5$ through $k=7$), with gains exceeding 15\% in solve rate. These results demonstrate the robustness and generalizability of our method across different model architectures and compositional difficulty levels.

\begin{figure}[t]
    \centering
    \includegraphics[width=0.9\columnwidth]{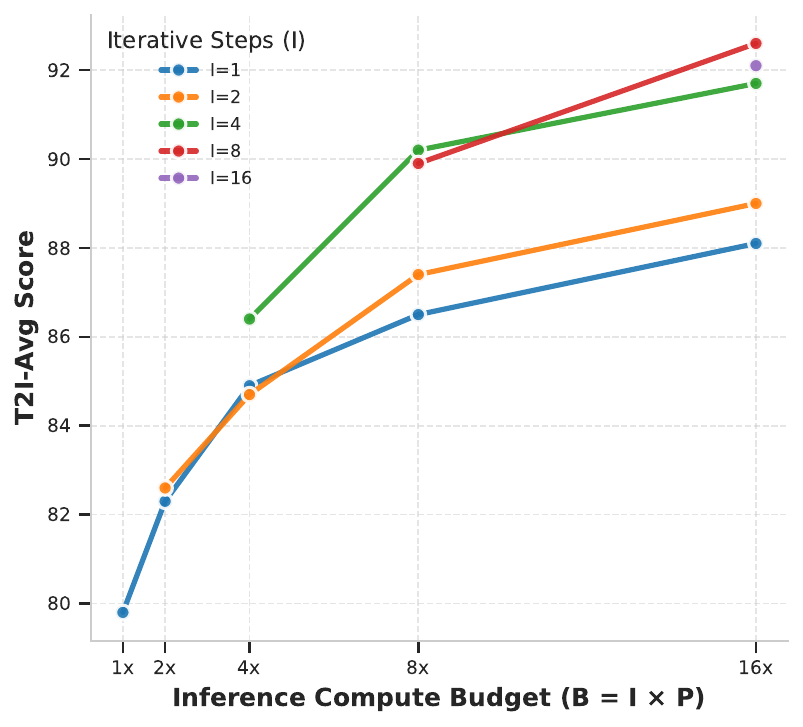}
    \caption{Comparison of iterative and parallel compute allocations on T2I-Bench.
Given a budget of $B=16$, mixed allocations of 8 iterative to 2 parallel generally outperform purely parallel or purely iterative strategies.}
    \label{fig:iter-par-comp-t2i}
\end{figure}

We show further results of our compute-budget allocation experiments for T2I Bench in Figure~\ref{fig:iter-par-comp-t2i}.  As shown, higher iterative compute is linked to higher T2I-Avg score, and  mixed allocations of 8 iterative to 2 parallel generally outperform purely parallel or purely iterative strategies. This follows a similar pattern  to conceptmix results reported in original paper.For these experiments, we used a subset of 30 prompts for each T2I-Bench category for compute reasons.

\end{document}